\documentclass{article}

\usepackage[preprint]{corl_2026} % Uncomment for pre-prints (e.g., arxiv); This is like ``final'', but will remove the CORL footnote.
\usepackage{booktabs}   % tables in results.tex
\usepackage{graphicx}   % figures in results.tex
\usepackage{siunitx}    % \SI{...}{...} in results.tex and method.tex
\usepackage{tikz}       % schematic in method.tex
\usetikzlibrary{positioning,arrows.meta,fit,backgrounds,calc}

\title{Teaching a Tree to Recover: Online Distillation of an LLM Supervisor with Counterfactual Replay}

\author{
  Kristian Ceder \qquad Kilian Freitag \qquad Knut Åkesson \\
  Department of Electrical Engineering, Chalmers University of Technology, Sweden \\
  \texttt{\{cederk, kilian.freitag, knut.akesson\}@chalmers.se}
}

\begin{document}
\maketitle

%===============================================================================

\begin{abstract}
Autonomous mobile robots commonly pair a global planner with a local model predictive controller (MPC), but nonconvex obstacle avoidance and a limited horizon can make the robot freeze in complex layouts. Deciding whether to replan, wait, or call for help is a supervisory decision. Recent work queries a large language model (LLM) supervisor on every anomaly, which is slow and costly. We instead distill the LLM online into a decision tree. After each training episode, counterfactual replay labels recoveries as needed, unnecessary, or harmful. These labels retire unhelpful rules. In simulation, on 20 unseen layouts where the MPC alone freezes, the tree handles at least 18 correctly across all seeds, either by reaching the goal or by calling for help when no feasible path exists. The tree matches both an always-query LLM supervisor and expert-designed rules, while issuing no LLM queries at test time. Handwritten rules cannot anticipate the long tail of situations a physical robot will encounter. Online distillation is a promising remedy: the tree handles familiar cases, and the LLM is consulted only for new ones, whose validated recoveries become new rules. Language models can thus also serve as designers of supervisors rather than only as runtime supervisors.

\end{abstract}


%===============================================================================

\begin{figure}[h]
\vspace{-5mm}
\centering
\includegraphics[width=\linewidth]{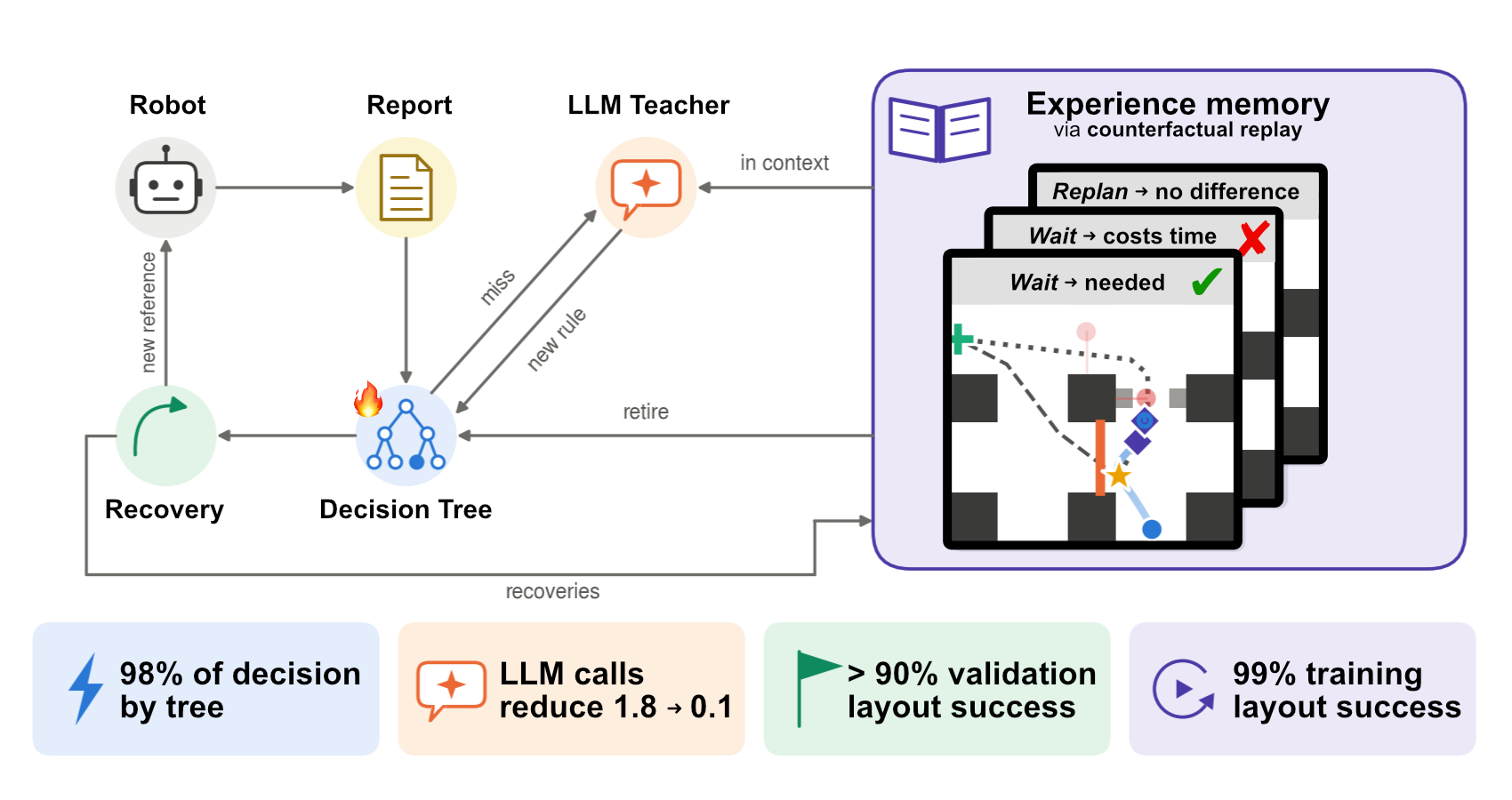}
\caption{An LLM teaches a decision tree to recover an MPC-driven robot, then leaves the loop. Left: the tree answers what it has seen, the teacher the rest. Right: the experience memory, built by counterfactual replay: the three recoveries of one learning layout, each re-run without itself and labeled needed, unnecessary (no difference) or harmful (cost time); the labels retire rules that only look right and are shown to the teacher in context. Bottom: 200 learning layouts, three runs.}
\label{fig:teaser}
\vspace{-5mm}
\end{figure}

%===============================================================================

\section{Introduction}
\label{sec:intro}

Autonomous Mobile Robot (AMR) navigation is commonly split into two layers. A global planner computes a path over a map, typically with Dijkstra's algorithm or A*, treating the robot as a point and ignoring its kinematic and dynamic constraints. A local controller, often a model predictive controller (MPC), tracks this path while respecting those constraints, and may deviate from it temporarily to avoid static and dynamic obstacles. Such deviations must stay small, because obstacle avoidance makes the optimization problem nonconvex: with a finite horizon, the solver is prone to local optima, slow or failed convergence, and exceeding its computation budget. When the path is blocked by a large or non-convex obstacle that is missing from the map, such as a blocking object across an aisle or a U-shaped arrangement of pallets, a way around it lies beyond what the controller can find, and the robot comes to a standstill. Whether it should then replan, wait, or call for help depends on the cause. This decision belongs to a supervisory layer above the controller. Navigation stacks typically implement this layer as hand-written recovery behaviors triggered when the planner or controller fails~\citep{macenski2020marathon}. In this work, we investigate how to learn this recovery behavior. Recent work places a large language model (LLM) in the loop as a supervisor~\citep{elhafsi2023semantic,sinha2024realtime,garg2024deadlock} and queries it on every anomaly, at a latency of seconds.

This paper studies a similar approach where an LLM is used to find effective failure recoveries for an MPC-controlled mobile robot based on our previous work~\citep{zhang2023case,ceder2024bev}. Unlike other approaches, we study where a decision should reside once it has been made a few times by abstracting and refining recurring events into a decision tree. Further, in subsequent loops, the decision tree is first queried, and only when no fitting decision is present the LLM is queried again for additional labels. Therefore, the cost of decisions drastically decreases over time. To refine and verify decisions in the tree, we use counterfactual replay of each action after an episode finishes, which enables in-context learning for the LLM supervisor. Through this, the LLM can continuously improve its decisions and update the decision tree.

In various simulated scenarios, we find that our method complement and assist the MPC controller, leading to a near-perfect success rate in our tests. Importantly, it performs on par with hand-designed rules specifically created for the tested scenarios, showing that our framework can adapt to and learn from observed scenarios without human supervision.

% Results
In 20 held-out test layouts where the MPC alone fails, we show that our method instead succeeds in reaching the goal position in more than 90\% of the cases. Furthermore, when comparing our method with a baseline of scripted rules specifically designed for those scenarios, we find that our method automatically finds equal or better decisions while reaching the goal in 10\% fewer steps, thus verifying its effectiveness at finding reasonable interventions.

Our contributions are:
\begin{itemize}
\item \textbf{Online distillation of an LLM for AMR failure recovery into a decision tree.} At a regular interval, a language model reads the current status report and decides whether an intervention is needed and picks a predefined recovery action, which acts only through the reference of an MPC tracker, the robot never waits for the LLM. To reduce the cost of repetitive interventions, we distill the LLM supervisor's actions online into a decision tree when they occur more than once.

\item \textbf{Counterfactual replay for in-context learning and rule-retirement.} We show that when having access to a simulator then re-simulating each situation without the supervisor's intervention gives labels if the action was needed, unnecessary, or harmful. Those are then used to retire existing rules and train the LLM in context to improve future decisions.

\end{itemize}

The code, data and videos are available at: \\ \url{https://github.com/kristianceder/Counterfactual-Tree-Teaching}.

%===============================================================================

\providecommand{\needcite}{\textcolor{red}{\textbf{[need citation?]}}}

\section{Related work}
\label{sec:related}

Closest to our work are systems that turn a foundation model's answers into reusable rules. \citet{peddi2023dtmpc} pair a decision tree with an MPC to diagnose actuator and sensor faults from failure data. Zetta~\citep{zetta2026} evolves runtime critics and recovery skills around a frozen manipulation policy. AgentChord~\citep{agentchord2026} pre-compiles recovery branches into a task graph. \citet{abaza2026semantic} promote vision-language resolutions of navigation instructions to deterministic rules between sessions. \citet{styrud2025betr} let a model add preconditions to a behavior tree when an action fails, to prevent it from happening again. Co-$\pi$-tree~\citep{copitree2026} distills a model's reasoning into a policy tree offline, and STAR~\citep{sakib2025star} stores diagnoses and recoveries in a growing knowledge graph. None of these add rules during an episode for a mobile robot among pedestrians, or evaluate whether the stored recoveries were needed. Counterfactual evaluation of an intervention has a precedent in Waymo's simulation of undisengaged drives~\citep{schwall2020waymo} and in Causal Agent Replay~\citep{shah2026car}, which intervenes on a single step of a software agent to find the exact step that caused the error. We pose a similar question, and leverage it as a learning signal for each recovery a supervisor orders. Lastly, \citet{sinha2024realtime}, whose fast classifier and slow reasoner over an MPC is the closest architecture, list learning from the slow reasoner's verdicts as future work, which is where we position our method.

%===============================================================================

\section{Method}
\label{sec:method}

The problem we address is a mobile robot that has to reach a goal in a warehouse. We consider approaches where a planner such as A* is first used to produces a route over the free space it knows about, and a controller then tracks a dense reference along it.
For instances, when using an MPC, as in our case, it solves a constrained optimization every iteration with obstacles as hard constraints to guarantee safety. Specifically, we address scenarios where the controller cannot independently find satisfactory solutions, such as the robot getting stuck due to obstructions unknown to the planner or the solver not finding a valid solution in time.
We base our tracker on the MPC part of our earlier hybrid controller ~\citep{zhang2023case,ceder2024bev}. The only part our method changes are two inputs that a supervisor needs: a \emph{hold} that pins the reference to the current pose and a heading target tracked during one.

\begin{figure}[t]
\centering
\includegraphics[width=\linewidth]{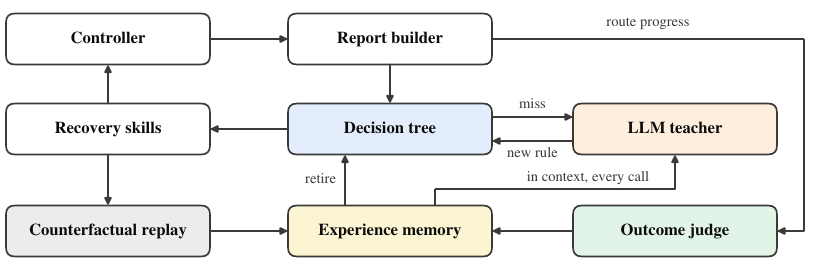}
\caption{The supervision loop. Top: the control loop at \SI{5}{Hz}, the same in every arrangement. Middle: every \SI{2}{s} the tree answers the report if a trusted rule matches and otherwise asks the model, whose answer becomes a rule. Bottom: between episodes each recovery is replayed with \textsc{continue} in its place and the label is filed on the rule that ordered it; retirement reads that record, and so does the teacher. A tree hit takes about \SI{6}{\micro\second}, a model call a median of \SI{1.5}{s}.}
\label{fig:loop}
\vspace{-5mm}
\end{figure}

We propose using a \textbf{supervision loop} on top of the existing control framwork as presented in Fig.~\ref{fig:loop}. This loop regularly receives a \textbf{report} that summarizes the current status of the controller and robot, including various \textbf{failure modes}. This \textbf{report} is then passed asynchronously to a \textbf{decision tree} that tries to classify which of the \textbf{recoveries} to trigger or if none is needed. If the \textbf{decision tree} does not return a result for the current situation, the \textbf{report} is instead passed to an \textbf{LLM teacher}. Decisions taken by the \textbf{LLM teacher} are saved and distilled into the \textbf{decision tree} if they happen more than once to continuously decrease the cost of querying recoveries. Further, we perform \textbf{counterfactual replay} of any taken decision to find out what would have happened if no action was taken instead. This is used to post-hoc verify the effect of a \textbf{recovery}, which is used for in-context learning of the \textbf{LLM teacher}. The details about all components are introduced next in their respective sections.

The scenarios we consider are layouts with staged situations and known solutions to make them verifiable. (Fig.~\ref{fig:scenario-grid}, App.~\ref{app:grid}) including temporary static blockages unknown to the planner such as parked vehicles, pedestrians passing across the planned route, a sealed passage, cases where no recovery can help reaching the goal because of constant blockage (only requesting human help solves this), and cases where no recovery is needed at all. An episode ends at the goal, on contact, when a human is called, on the watchdog, or when its step budget runs out, and \emph{ends right} if the robot reaches the goal on a layout with a way through or calls a human on one without.

\paragraph{Report.}
At each tick, the report builder records recent ego motion, the nearest static and dynamic obstacles and whether they block the route, controller deadline misses, the watchdog's allowance, the step budget, and what has been tried in the episode. These fields provide evidence, such as time stopped or distance to an obstacle, rather than verdicts such as \textsc{stuck}. Each decider then reaches its own verdict independently. Route progress is deliberately omitted in the report so that deciders cannot score themselves. Further, to identify a blocked route from the first tick, the static obstacle check uses the true map along the entire planned route.

\paragraph{Failure modes.}
We consider six failure modes, each opening with a decisive test on named fields of the report: \textsc{stuck} (stopped for seconds with no hold running), \textsc{oscillation} (repeated heading reversals without progress), \textsc{reverse\_tracking} (driving backwards for seconds), \textsc{collision} (a gap at or below zero), \textsc{timeout} (fewer steps left than the distance needs) and \textsc{solver\_deadline\_miss} (repeated deadline misses with no hold running). The used thresholds are in App.~\ref{app:details}.

\paragraph{Recoveries.}
Every recovery acts through the reference. \textsc{replan\_route} plans again over the full static map, which contains the obstruction the first plan did not; \textsc{wait} holds position until the obstacle it was ordered for has left the route or has stood still long enough to count as abandoned; \textsc{reorient} holds and turns to face along the route; \textsc{resume\_route} re-attaches the known route; \textsc{request\_human} ends the mission and is right on two grounds only, a replan that found no route or a blocker that has stood still after a \textsc{wait}; \textsc{continue} does nothing. A loop breaker bounds the supervisor: once several recoveries have passed without progress, it answers the tick itself with a replan, or a human call where its grounds are met (App.~\ref{app:details}).

\paragraph{The LLM teacher}

Leveraging the LLM's zero-shot capabilites for problemsolving are costly both in monetary terms and latency. The language model is asked only about the reports the tree cannot answer, which are by construction the situations no rule has covered yet. What makes a model useful there is that it can weigh a new report against a saved context rather than against thresholds alone: its prompt can describe the failure modes with their tests, the recoveries with when each applies, and a few consistency checks, such as that a running hold is not a failure. The model answers with a failure mode, or none, and an action.

\subsection{Distilling the teacher into a decision tree}
\label{sec:method-loop}

A report never repeats exactly, so the tree is keyed on a \emph{signature}: every piece of evidence the modes and recoveries are described by, binned at the thresholds the tests state, in the order the prompt checks them (App.~\ref{app:details}). Because the bin edges are the tests' own numbers, a learned rule reads like a hand-written threshold, and the tree can be printed and audited. A rule is a prefix of the signature and an answer, and the tree grows by contradiction rather than by construction: rules enter shallow, at least as deep as the evidence of the mode or action they name, and are deepened only when a teacher answer disagrees with the covering rule, which splits it at the first feature on which the two reports differ; an answer that agrees raises the rule's support. A rule is \emph{trusted} once the teacher has confirmed it in more than one episode and it has not been retired; a path that keeps being contradicted loses the tree's trust, and its reports always go to the teacher. A tick is resolved in order: the breaker if it applies, a trusted covering rule, otherwise the teacher.

\subsection{Experience memory, counterfactual replay and the outcome judge}

The method leverages two mechanisms to learn from experiences. The first, and the naive way to measure if a recovery action was successful is to measure progress toward the goal over time. In other words, did the action taken ensure progress for the agent? This signal is used by the outcome judge to classify online whether a recovery action was likely helpful.
However, progress alone cannot evaluate the necessity of an action, it might be equally good to not intervene at times. For that, we use counterfactual replay by re-simulating scenarios: Layouts and pedestrians are deterministic given the seed and every verdict is recorded with its step, so a recorded episode can be re-run with one verdict replaced by \textsc{continue}. Depending on the outcome, a recovery is labeled as \emph{needed} if the robot now fails or arrives clearly later, \emph{unnecessary} if it arrives about as soon, \emph{harmful} if it arrives clearly sooner or succeeds where the recording failed, or \emph{unclear} if both fail.

No matter which outcome, both signals are written into the \emph{Experience memory} where each row corresponds to a rule. Each row then containes information about the outcome judge's scores with a counter for the number of times it has fired, and a table of situation-action pairs (from counterfactual replay) with how each scenario ended, again with corresponding couters. While the judge's verdicts are written online, the replay is done offline.
Furthermore, the counterfactual replay is also used to retire rules that the outcome judge initially deemed helpful. This happens once 10 of its recoveries have replayed as unnecessary or harmful while at most one has replayed as needed, since harmful mostly means a later arrival and needed means a lost episode. A retired rule no longer answers: its situations go back to the teacher, which sees the retirement in its experience text, and its new answer replaces the rule.

%===============================================================================

\section{Experiments}
\label{sec:result}

\subsection{Experimental setup}
\label{sec:result-setup}

\paragraph{Environment and controller.}
The MPC and code is based on ~\citep{zhang2023case,ceder2024bev} which was built with OpEn~\citep{sopasakis2020open} and CasADi~\citep{andersson2019casadi}, and uses a horizon of \SI{4}{s} and a control period of \SI{200}{ms} (\SI{5}{Hz}). Each solve is capped by wall time, so the controller depends on the machine and its load (see App.~\ref{app:details}). The watchdog ends an episode once 45 of the last 60 solves have overrun the control period.

\paragraph{Supervisor and teacher.}
The robot is assessed every \SI{2}{s} from a report over the last \SI{5}{s} of state with about forty fields. As teacher, we use GPT-6 Luna through the provider's API with reasoning off and temperature 0 for deterministic answers.
A call carries about 5.8k tokens and answers in a median of \SI{1.5}{s}, and the calls of a 200-layout run cost under two US cents. The tree's, recoveries', judge's, replay's, and experience text's parameters are in App.~\ref{app:details}. A tree is evaluated on its own by running it \emph{frozen}, with no onboard model or learning. If no rule covers a report, the tree returns \textsc{continue}, and the report is counted as unanswered.

\paragraph{Scripted rules.}
The hand-written baseline is the supervision loop of Sec.~\ref{sec:method} with the failure tests as its decider: at each tick it applies the six tests to the report literally, in the order the prompt checks them, and answers with the fixed recovery of the mode whose test holds, or \textsc{continue} when none does. It shares the report, the recoveries and the breaker with every other arrangement, so a comparison against it measures the decider in isolation.

\paragraph{Layouts, deciders and runs.}
The used layouts are a $3{\times}3$ block world of aisles, so all experiments use the scenario grid seen in (App. \ref{app:grid}): 200 layouts (seeds 100--299) for learning, 20 held-out layouts (seeds 300--319) and 100 unseen ones (seeds 400--499), of which 49, 2 and 28 need a human. Four deciders share the report and recovery library: \emph{no recovery}; the \emph{scripted rules}; the \emph{LLM alone}, asked every tick; and the \emph{frozen tree}, grown over the 200 learning layouts by the method and run with no model. The method, the two ablations and the LLM alone were each run three times, the scripted rules and no recovery, which call no model, once.

\subsection{Held-out layout evaluation}
\label{sec:result-main}

\begin{table}[t]
\centering
\caption{Held-out layout mean over three runs ($\pm$~s.d.). \emph{Right}: layouts ending right, of 20 (the goal, or a human called where one was needed); \emph{T\,/\,W\,/\,C}: timeouts / watchdog stops / collisions; \emph{steps}: to the goal; \emph{age}: control cycles from report to verdict; \emph{tree}: ticks answered by a rule. Frozen tree: the tree each method run ended with after the 200 learning layouts, run with no model.}
\label{tab:main}
\small
\begin{tabular}{lcccccc}
\toprule
Arrangement    & Right /20      & T\,/\,W\,/\,C & Steps        & Calls / ep     & Age           & Tree  \\
\midrule
No recovery    & 0              & 14 / 6 / 0    & --           & --             & --            & --    \\
Scripted rules & 18             & 2 / 0 / 0     & 288          & --             & 1.0           & --    \\
LLM alone      & $16.3 \pm 1.5$ & 3.3 / 0 / 0.3 & $259 \pm 20$ & $24.6 \pm 3.3$ & $7.8 \pm 1.0$ & --    \\
Frozen tree    & $18.3 \pm 0.6$ & 1.7 / 0 / 0   & $260 \pm 5$  & 0              & 1.0           & 99\%  \\
\bottomrule
\end{tabular}
\end{table}

Without a recovery layer, the controller ends no held-out layout right (Table~\ref{tab:main}), it stalls at the wall until the step budget runs out or the watchdog stops it. The scripted rules instead end 18 of 20 correctly. When querying an LLM every tick, it ends 16, 15, and 18 correctly for each seed, respectively, with about 25 calls per episode and verdicts acting eight cycles after their report. Further, the frozen tree of each method run succeeds in 18, 19, and 18 cases with no LLM at all during evaluation. Thus, the frozen tree performs on par with the rules and on average outpeforms the LLM alone that it was distilled from. It answers 99\% of ticks from an existing rule, and it reaches the goal a tenth sooner than the rules because its wall rule replans before the robot has stopped. On the 100 unseen layouts, the frozen trees end 96, 92 and 96 of the scenarios successfully against the rules' 95, and ends 17--27 steps sooner when both reach the goal.

\subsection{Learning evaluations}
\label{sec:result-method}

\begin{table}[t]
\centering
\caption{The 200 learning layouts, the same seeds for every arm; means over three runs, $\pm$~s.d.\ (the scripted rules were run once). \emph{T\,/\,W\,/\,C}: timeouts / watchdog stops / collisions; \emph{recov.}: recoveries ordered, of which \emph{waits}; \emph{needed}: share the replay labeled needed, where it runs; \emph{rules (ret.)}: final rules (retired by replay labels); \emph{unseen}: the tree each run ended with, run frozen on the 100 unseen layouts, right of 100 (rules: on the same layouts).}
\label{tab:method}
\footnotesize
\setlength{\tabcolsep}{3.5pt}
\begin{tabular}{lccccccc}
\toprule
Arm                & Right /200               & T\,/\,W\,/\,C & Recov.       & Waits        & Needed & Rules (ret.)   & Unseen /100             \\
\midrule
Scripted rules     & 197                      & 2 / 0 / 1     & 353          & 148          & 67\%   & --             & 95                      \\
\midrule
Distillation alone & $192 \pm 0$              & 7 / 0 / 1     & $673 \pm 0$  & $517 \pm 0$  & 37\%   & $8 \pm 0$ (0)  & $92 \pm 0$              \\
Without replay     & $194.0 \pm 1.7$          & 5.0 / 0 / 1   & $679 \pm 5$  & $505 \pm 11$ & --     & $15 \pm 6$ (0) & $92.7 \pm 0.6$          \\
Method             & $\mathbf{198.0 \pm 1.0}$ & 2.0 / 0 / 0   & $552 \pm 46$ & $351 \pm 31$ & 41\%   & $30 \pm 5$ (3) & $\mathbf{94.7 \pm 2.3}$ \\
\bottomrule
\end{tabular}
\end{table}

\begin{figure}[t]
\centering
\includegraphics[width=0.87\linewidth]{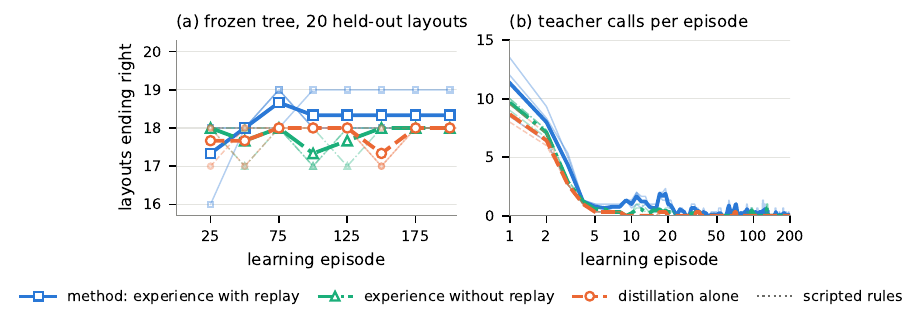}
\caption{The 200 learning layouts in sequence: thin lines are runs, thick their mean. (a) Every 25 episodes, the tree is saved and run frozen, with no model, on the 20 held-out layouts: layouts ending right, of 20 (dotted: the scripted rules' 18). (b) Teacher calls per episode, a 3-episode moving average on a logarithmic episode axis.}
\label{fig:heldout-curve}
\vspace{-5mm}
\end{figure}

\emph{Only the method keeps the rules' level while learning.} The scripted rules end 197 of the 200 learning layouts right (Table~\ref{tab:method}) and the method 199, 197, and 198, respectively. While the scripted rules timeout in 2 scenarios and crash in 1, our method reaches 1--3 timeouts in different layouts, thus leading to comparable performance. All of the failed scenarios involve a wall and gate, where conservative behaviors delay progress.
Distillation alone ends 192 in all three runs, failing on the same eight layouts, and experience without replay 192, 195 and 195. Paired run by run, the method is right on 5--7 layouts that distillation is not and never the other way, and on 3--7 against 0--1 for experience without replay. Every learned tree reaches the goal in 24--33 steps before the rules, since each learned the early wall replan. Therefore, the difference in the ablations stems from other layouts. 

\emph{Importance of counterfactual replay and retirement.} Every tree without counterfactual replay and retirement keeps the rule the teacher first gives, for instance, for a pedestrian on the route, ``not holding, moving, dynamic obstacle ahead: \textsc{wait}''. The judge scores this rule good almost every time, since a robot held behind a pedestrian who walks on does get going again. These trees wait behind every pedestrian, 498-517 times per run against the rules' 148, and lose layouts of two kinds: breakdowns, where the blocker never walks on, and the renewed waits run the episode out instead of calling a human, and layouts that add gates to a wall or a reversed start, where the waits use up the step budget. In all three method runs, the replay retires that rule at the sixth episode, after ten of its waits came back unnecessary or harmful and it was never labeled needed. This example shows that both counterfactual replay and retirement are key components of the method.

\emph{Learning remains stable on unseen and held-out layouts.} Every checkpoint of every run ends 16--19 of the 20 held-out layouts correctly (Fig.~\ref{fig:heldout-curve}a), showing that the core method is quite robust and does not overfit. At the end of training, our method ends correctly on 18.3 of 20 held-out layouts, while all others end at 18. This is due to a single seed finding 19 solutions correctly.
On the 100 unseen layouts, the method's trees end 96, 92, and 96 correctly against 92 for every distillation run and 93, 93, and 92 without replay. The scripted rules reach 95 correct layouts, thus outperforming our method in one case (See Table~\ref{tab:method}).
Meanwhile, the LLM gradually leaves the loop (Fig.~\ref{fig:heldout-curve}b): the method asks the teacher 1.8 times per episode over the first 25 layouts and 0.1 over the last 25 (1.5-2.0 and 0-0.2 across runs), 73-81 calls in all, and its tree answers 98\% of ticks.

\subsection{Self-improvement on a repeated layout}
\label{sec:result-permap}

\begin{figure}[t]
\centering
\includegraphics[width=\linewidth]{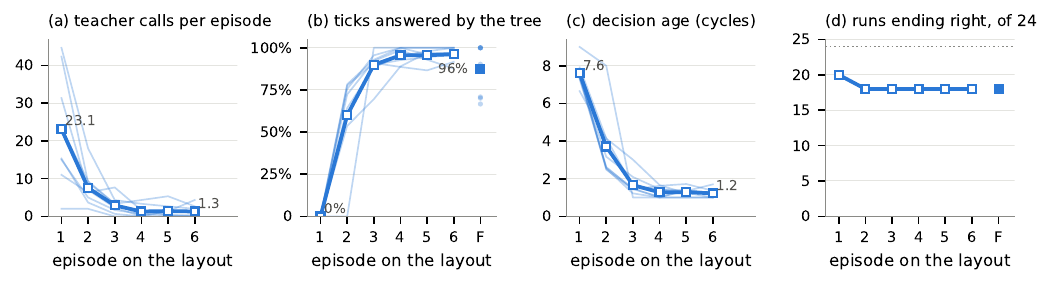}
\caption{Self-improvement on a repeated layout. For each of eight layouts, the method starts from an empty tree and drives the same layout six times in a row; the tree it has after the sixth is then frozen, with no model, and drives it once more (F). Three independent repeats per layout, each from an empty tree; thin lines are layouts (mean of their repeats), thick lines are their mean. (a) Teacher calls per episode. (b) Ticks answered by a rule. (c) Cycles from report to verdict. (d) Runs ending right, of the 24 (eight layouts, three repeats).}
\label{fig:permap}
\vspace{-5mm}
\end{figure}

In our learning stream, no layout gets repeated. In this ablation, we test revisiting one repeatedly from an empty tree, which shows the model leaving the loop directly (Fig.~\ref{fig:permap}): the teacher answers 23 ticks per episode on the first visit, 7.5 on the second, and about 1.3 from the fourth on, when the tree answers 96\% of ticks and a verdict acts 1.2 control cycles after its report instead of 7.6. The endings do not change: 20 of the 24 runs end right on the first visit and 18 on every later one, frozen included. The failures are the three layouts with gates, where the tree waits and resumes behind pedestrians until the step budget runs out, as discussed above. The replay then labels those recoveries harmful 179 times against 17 needed and retires the rule in 8 of the 9 runs, but the teacher, asked again about the same situation, tends to give the same answer, and in two runs re-teaches the retired rule at the next visit.
This uncovers a weakness of the method, indicating that the LLM teacher is not sufficiently considering the in-context experience memory when introducing rules.

\subsection{Counterfactual necessity of the recoveries}
\label{sec:result-counterfactual}

In most recoveries, the learned trees order is not needed (Table~\ref{tab:counterfactual}, App.~\ref{app:details}): of those whose replay was clean (98--99\%), the replay labels 37\% of distillation's and 41\% of the method's needed. The largest share, 46 and 41\%, is harmful only in the weak sense that the robot would have arrived more than 10 steps sooner without it, and almost all of it is waits behind pedestrians who would have cleared the route anyway. Harmful in the strong sense, i.e., a replay that turns a wrong ending right, is 6\% for distillation alone and 1\% for the method. The scripted rules' recoveries, replayed the same way, are needed 67\% of the time: they act only once a failure test has fired, so their waits are needed three times in four against the trees' one in five, and their human calls always. The replay's value to the loop is therefore the few retirements it makes of rules the judge scores well.

%===============================================================================

\section{Discussion}
\label{sec:discussion}

The learned tree's performance suggests that an LLM can teach a compact, fast supervisor during deployment. Matching the scripted rules is meaningful here because the tree was not given their decision logic: it acquired its own conditions from the teacher's responses. While it had structured information that was targeted to the specific problem, such as reports and available recoveries, it successfully discover itself when and how to best use them. Further, the distillation follows the expected trend of decreasing the teacher calls over training and thus substantially decreases the overall cost of predicting repetitive recoveries.

The pedestrian case illustrates why successful outcomes alone are a weak learning signal. Waiting can appear effective because a moving pedestrian eventually clears the route, even when waiting was unnecessary. A judge that sees only the eventual progress reinforces this rule, and the error becomes apparent only when the same rule is applied to a persistent blockage or a time-constrained layout. Counterfactual replay gives the supervisor a way to question an apparently successful action and retire the rule that produced it. The repeated-layout experiment also shows the limit of this mechanism: after retirement, the teacher can propose the same rule again. Preventing that cycle may require the teacher to account more directly for why an earlier rule was rejected.

Future work will investigate whether these findings carry over to a real robot, particularly whether the sim-to-real gap makes counterfactual replay less reliable. Additionally, further experiments that vary rule retirement and the judge's verdict separately could establish how much each contributes to the observed results.

\section{Conclusion}
\label{sec:conclusion}
We distilled a language model supervising an MPC tracker through its reference into a decision tree during deployment for failure recovery. On staged situations with known answers, the final distilled decision tree, run without an LLM, gets 18--19 of 20 held-out layouts right, matching hand-written rules while reaching the goal about a tenth sooner by preemptively detecting failures. Further, we show the importance of post-hoc evaluating decisions to consequtively refine and prune the tree if consequences were different than anticipated by the teacher. Overall, our findings reinforce the substantial potential of using automatic teachers in combination with distillation as high-level reasoners in robotic failure discovery.

%===============================================================================

\clearpage
% The acknowledgments are automatically included only in the final and preprint versions of the paper.
\acknowledgments{This work is supported
by the ITEA project ArtWork and the Vinnova project DrivePR.}
%If a paper is accepted, the final camera-ready version will (and probably should) include acknowledgments. All acknowledgments go at the end of the paper, including thanks to reviewers who gave useful comments, to colleagues who contributed to the ideas, and to funding agencies and corporate sponsors that provided financial support.
%===============================================================================

\bibliography{example}  % .bib

\appendix

\section{The scenario grid}
\label{app:grid}

\begin{figure}[h]
\centering
\includegraphics[width=\linewidth]{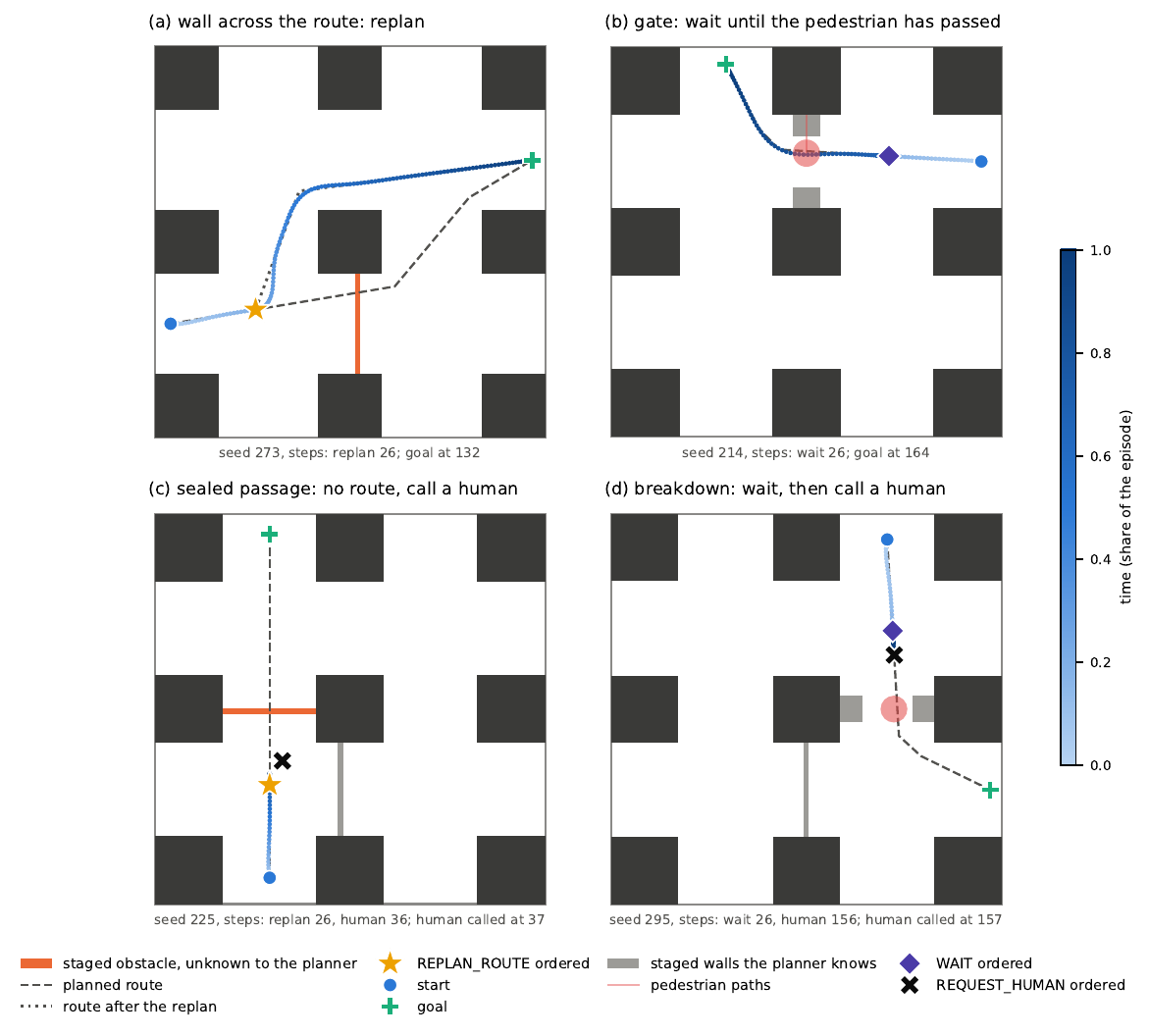}
\caption{The scenario grid, one staged situation per panel, from late in the method's first learning run, where the tree answers every tick. Dark: the $3{\times}3$ blocks; orange: a staged obstacle the planned route (dashed) runs through, which the planner was not told about; gray: staged walls the planner knows, which form a gate or close the other way round. The driven path is shaded by time; red: pedestrian paths, and each pedestrian's position at the first recovery. (a) A wall: the tree replans at the first tick, before the robot has stopped, and the new route (dotted) goes around the block. (b) A gate: the robot holds while a pedestrian stands in it, until the pedestrian has left. (c) A sealed passage: the replan finds no route, and the tree calls a human. (d) A breakdown: the blocker never leaves; the hold is let go once it has stood \SI{30}{s}, and the tree calls a human.}
\label{fig:scenario-grid}
\end{figure}

Every layout is the same block world, assembled from a seed by the scenario generator. The learning, held-out and unseen layouts come from its mixed stream, so one layout can hold several situations, a wall and a gate for instance, as well as patrolling pedestrians; 49 of the 200 learning layouts, 2 of the 20 held-out layouts and 28 of the 100 unseen layouts have no way through, of which 26, 2 and 21 are breakdowns and the rest sealed passages. The four panels are layouts that stage a single situation.

\section{Implementation details}
\label{app:details}

This appendix collects details of the method that Sec.~\ref{sec:method} states only in summary.

\paragraph{Controller.}
The tracker is that of our earlier hybrid controller~\citep{zhang2023case,ceder2024bev}, with changes made for this work of which three matter to the supervisor: a hold input that pins the reference to the current pose at zero reference speed; a heading target tracked during a hold, with the heading-deviation cost a hold relies on to turn in place; and a hard wall-clock limit of \SI{500}{ms} per solve was added compared to the previous \SI{5}{s}. Obstacles are hard constraints rather than penalties, so a robot pressed against a wall it was not told about stalls instead of cutting into it; dynamic obstacles are predicted at constant velocity. The tracker's lower reference-speed setting is used until the first successful replan and the higher one afterward. Because the limit is wall time, what the controller does depends on the machine and its load, not only its timing: every run reported here is on one laptop (Intel Core i7-13700H), otherwise idle, with the solver built for the \SI{500}{ms} limit. The supervisor assesses the robot every \SI{2}{s} from step 25 on; a tree hit answers in about \SI{6}{\micro\second} and a model call in a median of \SI{1.5}{s} (the method's runs; 1.2--\SI{1.6}{s} for the model alone).

\paragraph{The hosted teacher.}
GPT-6 Luna is called through the provider's responses API with reasoning effort ``none'' and temperature 0, from a worker thread. The compiled solver holds Python's interpreter lock for the whole of a solve, which starves a local model generating tokens in a thread; a call to a hosted model needs the lock only to send the request and read the reply, so a thread is enough. The fixed head of the prompt carries an explicit cache breakpoint; the experience text and the report follow it and are paid for in full. A failed call is retried by the client and then up to four more times \SI{20}{s} apart; none of the calls in the runs reported here failed. A method run makes 73--81 calls at about 5.8k prompt tokens each, and the model alone about 25 per episode.

\paragraph{Report and recoveries.}
The report also contains the robot's position, clearance by direction, whether a hold is active and how many recovery cycles have passed without progress. It contains evidence rather than verdicts with one exception, the supervisor's own earlier verdicts in the episode history. \textsc{replan\_route} plans through the visibility graph~\citep{lozanoperez1979visibility} of the full static map; it moves a start point that lies inside an inflation margin to the nearest free point within \SI{1}{m} and, if the new route leaves more than \ang{90} off the heading, turns first. \textsc{reorient} holds for up to 40 cycles. A \textsc{wait} kept past its 25 cycles is released once the obstacle has gone or has stood for \SI{30}{s}, after which the supervisor must be shown a robot that is not holding, so that it can call a human. Verdicts are edge-triggered, so a condition that holds for forty cycles is one failure, and a verdict that repeats a running recovery is a \textsc{continue}. The imminent-collision mode and its back-off recovery exist in the code and are switched off in every run reported here: a supervisor on a two-second clock with a verdict age of several control cycles cannot act on a pedestrian who closes within one to two seconds, and avoiding pedestrians is left to the MPC. The breaker fires once three recoveries have passed without a meter of progress; holds against a dynamic obstacle that has stood for under \SI{30}{s} do not count toward it, since three holds at a gate that stays shut are correct behavior, and it never fires into a running hold. Every \textsc{replan\_route} resets its count, on the reasoning that a new route is a fresh start; that includes a replan that found no route, which is how a loop of replans at a sealed passage stays below it. Two defects we have not evaluated remain: that reset, and a hold kept until clear, which is unbounded against an obstacle that keeps moving on the route.

\paragraph{Failure tests.}
The decisive tests the modes of Sec.~\ref{sec:method} open with: \textsc{stuck}, stopped $\geq$\SI{4}{s} with no hold running; \textsc{oscillation}, $\geq 4$ heading reversals with $<$\SI{0.2}{m} of progress in \SI{5}{s}; \textsc{reverse\_tracking}, reversing $\geq$\SI{3}{s}; \textsc{collision}, a gap at or below zero; \textsc{timeout}, fewer steps left than the distance needs; \textsc{solver\_deadline\_miss}, $\geq 10$ misses in \SI{12}{s} with no hold running. The prompt's watchdog check reads: a high miss count on an open road is \textsc{continue} while the watchdog has more than 20 misses left, \textsc{wait} at or below.

\paragraph{Tree.}
The signature has thirteen features; a rule enters on the first three, is trusted after the teacher has confirmed it in two distinct episodes, and a path is distrusted after three contradictions or three re-teachings. The bins are, in the order the tree may split on them (\texttt{tree.FEATURES}): whether a hold is active; time stopped (none, brief, $\geq$\SI{4}{s}); what blocks the route (static, dynamic, clear); whether a route to the goal exists, placed straight after the blockage it qualifies, since the same wall is a replan while a route exists and a human call once none does; contact or near contact; whether a dynamic obstacle is approaching; time reversing; heading reversals; progress over the window (back, stalled, forward); heading error to the route (on, off, away); whether the blocking obstacle is moving, parked or abandoned; solver deadline misses (ok, miss, or critical when the watchdog has 20 or fewer misses left on a clear road); and whether the remaining step budget is tight. The bin edges are the numbers stated in the failure modes' tests, with three added that no test uses (\SI{1}{s} for a brief stop or reversal, and \ang{30} and \ang{90} of heading error), and an absent field is its own bin. A teacher answer that disagrees with a rule on an identical signature is recorded as a contradiction and the newer answer wins. A rule whose named mode fails its own test on the report it was learned from is stored and flagged \emph{unevidenced}, never overruled; the depth floor applies however an answer arrives, including when a retired or contradicted rule is replaced.

\paragraph{Prompt.}
The prompt states the failure modes, the checks and the recoveries in prose. The complete system prompt, one report and the experience text of one escalated tick are released with the code.

\paragraph{Judge.}
Every recovery a rule stands for opens its window of 25 control cycles once its hold ends, and is credited on \SI{1}{m} of route progress within it. A recovery superseded before its window closes is settled when the replacement is ordered, good only if the meter was already gained. An episode ending inside a window settles it: the goal is good, a collision or a watchdog stop is bad. A chain of \textsc{wait}s is judged once, where it ends: \textsc{resume\_route} and \textsc{request\_human} continue a pending \textsc{wait} rather than superseding it. A replan that finds no route is left unjudged, and \textsc{request\_human} is judged on whether the layout needed one. A rule that answers \textsc{continue} is scored on the spot, against whether the report already met a failure test. A retired rule is kept in the tree as retired.

\paragraph{Experience text.}
The text shows the covering rule and up to eight nearest ones, and the scored situation--action pairs of the four most similar situations. Labels from drifted replays and unclear labels are ignored, and the breaker's own orders are never replayed. In the rules block the nearest rules are chosen by the same hold state and the most conditions in common with the report. The table gives, for each situation, how often each action was taken, how the judge scored it and how the replays labeled it; each situation names the features on which it differs from the current report and marks a feature on which the answer has differed before; the guide adds that a thin or mixed record leaves the decision to the instructions. The text is inserted after the cached head and before the report, so the cached prefix is unaffected, and it acts on first occurrences, confirmations and retired rules, which are the ticks on which the model is asked.

\paragraph{Counterfactual replay.}
``Clearly later'' and ``clearly sooner'' in the labels mean more than 10 steps. The replay runs between episodes on the CPU, once per recovery of the episode just driven, at most six, and a recovery rule is retired once ten of its recoveries have come back unnecessary or harmful while at most one label in ten says needed. Ticks after the replaced verdict are answered \textsc{continue} for 50 steps and by the scripted rules after that. For a recovery that is followed by another within those 50 steps the later one is removed as well, so \emph{needed} is an upper bound there. The replay checks itself against the recording frame by frame; the divergence, where there is one, is the controller's, not the supervisor's, since a solve that sits on the \SI{500}{ms} wall-clock cap returns a different control depending on scheduling noise. On the idle laptop 98--99\% of the learning arms' recoveries replayed exactly up to their own step; labels are counted only where the replay was clean up to the recovery's own step, and every counterfactual share is approximate. An iteration-capped solver would make the replay exact and is the change we would make first.

%TODO fix so it fits.
Note that it cannot score a \textsc{wait} ordered at a standstill and since a solve near its wall-clock cap returns a different control under scheduling noise~\citep{chance2022determinism}, the replay checks itself frame by frame against the recording and drops the label of a recovery whose replay drifted, so every counterfactual share in Sec.~\ref{sec:result-counterfactual} is approximate.

\begin{table}[h]
\centering
\caption{Counterfactual labels of the recoveries on the learning streams (Sec.~\ref{sec:result-counterfactual}), each replaced by \textsc{continue} in a replay run by the loop itself (at most six per episode); means~$\pm$~s.d.\ over three runs. \emph{Clean}: replays exact up to the recovery's own step, of which the shares are taken. \emph{Harmful}: the replay flipped the ending to right, or only arrived more than 10 steps sooner. Experience without replay runs no replay. The scripted rules' single run was replayed after the fact with the same settings (160 of 200 episodes exact; 94\% of recoveries clean to their own step).}
\label{tab:counterfactual}
\small
\begin{tabular}{lccccccc}
\toprule
                   &              &       &              &              & \multicolumn{2}{c}{Harmful} &         \\
\cmidrule(lr){6-7}
Arm                & Replayed     & Clean & Needed       & Unnec.       & ending      & steps         & Unclear \\
\midrule
Scripted rules     & 351          & 94\%  & 67\%         & 26\%         & 0\%         & 6\%           & 1\%     \\
\midrule
Distillation alone & $571 \pm 6$  & 98\%  & $37 \pm 1$\% & $10 \pm 0$\% & $6 \pm 0$\% & $46 \pm 0$\%  & 1\%     \\
Method             & $528 \pm 29$ & 99\%  & $41 \pm 1$\% & $17 \pm 3$\% & $1 \pm 1$\% & $41 \pm 4$\%  & 0\%     \\
\bottomrule
\end{tabular}
\end{table}

\end{document}